\documentclass[runningheads]{llncs}

\usepackage{eccv}

\usepackage[utf8]{inputenc}
\usepackage{textgreek}

\usepackage{eccvabbrv}

\usepackage{graphicx}
\usepackage{booktabs}

\usepackage[accsupp]{axessibility}  

\usepackage[hidelinks]{hyperref}

\usepackage{orcidlink}
\usepackage{marvosym}
\usepackage{booktabs}
\usepackage{multirow}
\usepackage{siunitx}
\usepackage{booktabs}
\usepackage{array}
\usepackage{placeins}
\usepackage{wrapfig}

\newcommand{\corrauth}{\texorpdfstring{\textsuperscript{\Letter}}{}}

\begin{document}

\title{Rethinking IRSTD: Single-Point Supervision Guided Encoder-Only Framework Is Enough for Infrared Small Target Detection} 

\titlerunning{SPIRE}

\author{Rixiang Ni\inst{1}\orcidlink{0009-0007-4614-9977}, Jun Chen\inst{1}\corrauth, Boyang Li\inst{1}\corrauth, Yonghao Li\inst{1}, Wujiao He\inst{1}, Yuji Wang\inst{1}, Feiyu Ren\inst{1}, Haoyang Yuan\inst{1}, Wei An\inst{1}}

\authorrunning{R. Ni. et al.}

\institute{National University of Defense Technology, Changsha 410003, China\\
\email{\{nirixiang, chenjun11, liboyang20\}@nudt.edu.cn}\\
Code and benchmark: \url{https://github.com/NIRIXIANG/SPIRE-IRSTD}
}

\maketitle
\begingroup
\renewcommand{\thefootnote}{\Letter}
\footnotetext{Corresponding authors.}
\endgroup

\begin{abstract}
	Infrared small target detection (IRSTD) aims to localize small targets in cluttered infrared scenes. Extensive research follows the pixel-level supervision-guided `encoder--decoder' segmentation paradigm. Although these methods have achieved promising performance, they often overlook that infrared small targets occupy only a few pixels and are usually surrounded by blurred, low-contrast boundaries caused by cluttered backgrounds. Based on this observation, we argue that the first principle of IRSTD should be target localization rather than dense reconstruction of target regions entangled with indistinguishable background noise. In this paper, we reformulate IRSTD as a centroid regression task and propose a novel Single-Point Supervision guided Infrared Probabilistic Response Encoding method (namely, SPIRE),
     which is non-trivial because point-level supervision must produce detection outputs comparable to dense supervision. Specifically, we design Point-Response Prior Supervision (PRPS) to transform single-point annotations into probabilistic response maps consistent with infrared point-target response characteristics, and combine it with a High-Resolution Probabilistic Encoder (HRPE) that performs encoder-only probabilistic regression followed by lightweight peak-based inference, without decoder reconstruction. By preserving high-resolution features and increasing effective supervision density, SPIRE alleviates optimization instability under sparse target distributions. Finally, extensive experiments on two public IRSTD benchmarks, SIRST-UAVB and SIRST4, demonstrate that SPIRE achieves competitive target-level detection performance with consistently low false alarm rate (Fa) and significantly reduced computational cost.
	
\end{abstract}

\section{Introduction}
\label{sec:intro}
Infrared Small Target Detection (IRSTD) is a fundamental component of infrared search and tracking systems \cite{2023LESPS}, with wide applications in traffic monitoring \cite{2023DNANet29,zhang2022isnet68}, early warning \cite{li2018robust30,xu2023multiscale59}, and maritime rescue \cite{zhang2022exploring67,zhao2024infrared72}.
Despite its practical importance, most existing IRSTD approaches are formulated under dense prediction paradigms, typically adopting encoder–decoder segmentation architectures trained with pixel-level annotations \cite{2023DNANet29,2023LESPS,2023MCLC,2023UIUNet,2024ISPANet,2024MSHNet,2024STTrans,2025FromEasy2Hard,2025HDNet,2025L2SKNet,2025SIRST-UAVB_Dataset,2026FeedBack}.

\begin{figure}[tb]
	\centering
	\includegraphics[width=\textwidth]{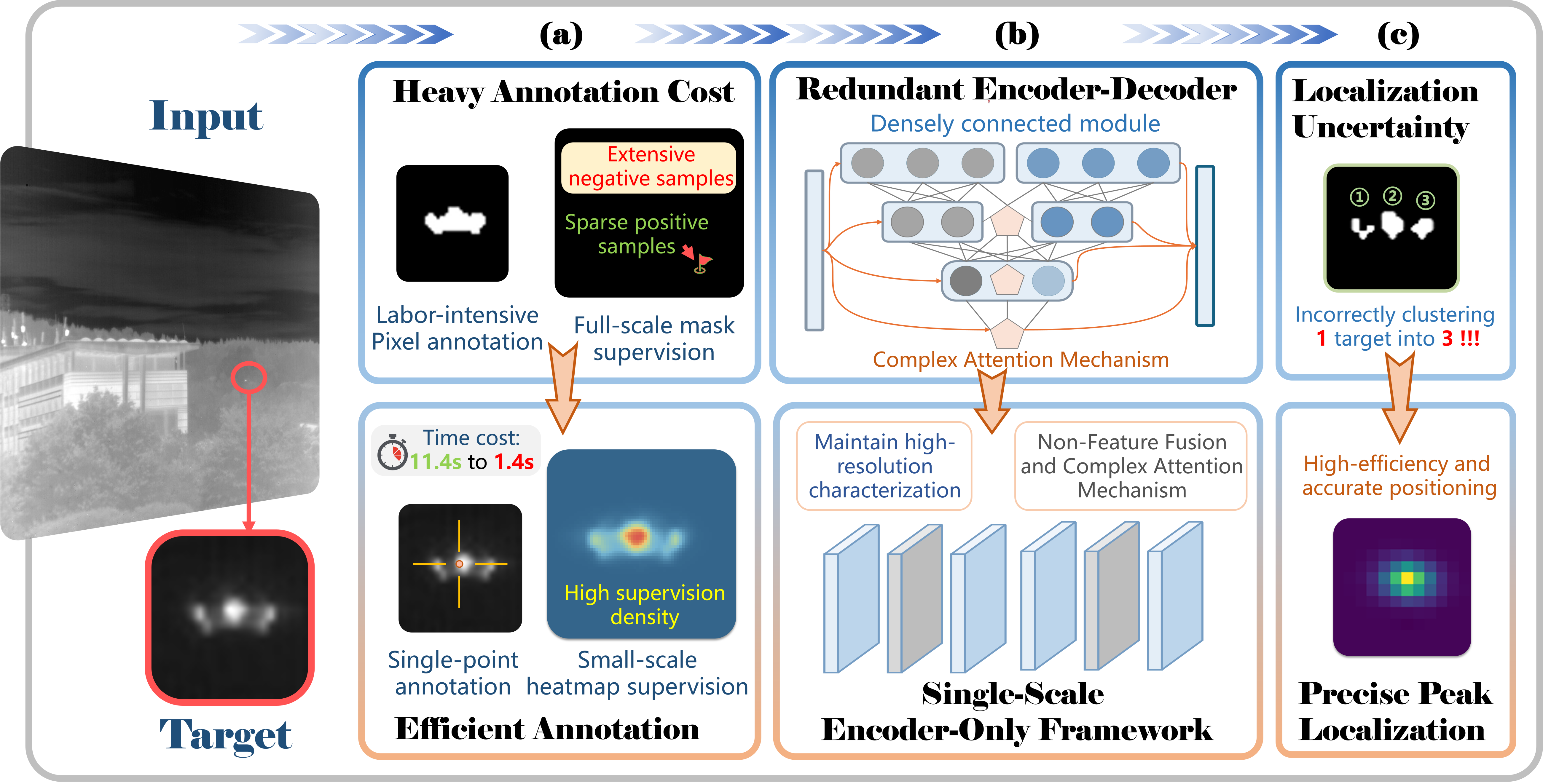}
	\caption{
\textbf{Motivation of SPIRE for IRSTD.}
Existing IRSTD methods suffer from three key limitations:
(a) heavy pixel-level annotation cost with extremely sparse positive samples,
(b) redundant encoder–decoder architectures relying on extensive feature fusion and attention mechanisms,
and (c) localization uncertainty caused by segmentation-based predictions.
SPIRE reformulates IRSTD as centroid localization with single-point supervision and probabilistic response map modeling, enabling efficient annotation, a lightweight encoder-only framework, and precise peak-based localization.
}
	\label{3problem}
\end{figure}

\textbf{Three limitations of the existing fully supervised encoder--decoder segmentation paradigm.} Although this paradigm has achieved promising performance, it introduces three inherent limitations.
\textit{(1) Heavy Annotation Cost.} Pixel-wise labeling is labor-intensive and subject to boundary and scale ambiguities (see \cref{3problem}(a)). \cite{2023MCLC} reports that single-point annotation reduces labeling time by 87.72$\%$, from 11.4 s to 1.4 s per image compared with pixel-wise annotation.
\textit{(2) Redundant Encoder–Decoder Framework.} To compensate for extreme foreground sparsity and stabilize optimization, multi-scale fusion and densely connected modules (e.g., nested structures, attention mechanisms) are widely employed, substantially increasing training and inference complexity (see \cref{3problem}(c)).
\textit{(3) Localization Uncertainty.} Instance-level localization is typically obtained through additional post-processing of segmentation masks, introducing pipeline overhead and potential error accumulation (see \cref{3problem}(c)).

\begin{figure}[tb]
	\centering
	\includegraphics[width=\textwidth]{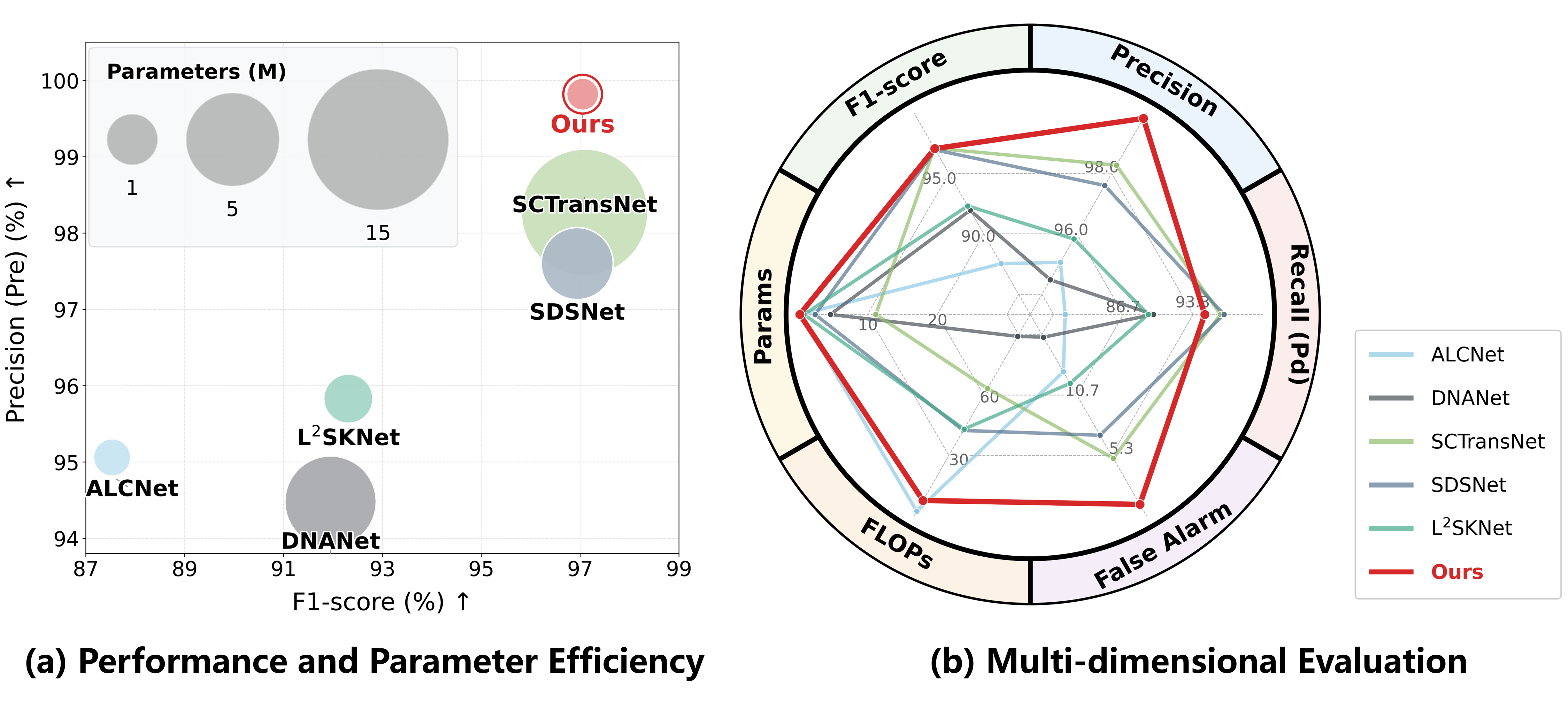}
	\caption{
\textbf{Performance of SPIRE on SIRST-UAVB.}
(a) Performance–efficiency trade-off among representative IRSTD methods, where the axes denote F1-score and Precision, and bubble size indicates the number of parameters.
(b) Six-dimensional evaluation across Precision, Recall (Pd), F1-score, False Alarm rate, FLOPs, and parameters.
SPIRE achieves a favorable balance between detection accuracy and computational efficiency.
}
	\label{fig:f1_pre_params_bubble}
\end{figure}

\textbf{The first principle of IRSTD \& two trade-off solutions.} Infrared small targets (IRSTs) are characterized by extremely limited spatial extent, Gaussian-like energy diffusion, low signal-to-noise ratios, and highly sparse spatial distributions. Considering the practical IRST search and tracking systems  primarily require reliable localization rather than contour reconstruction, \textit{we argue that the first principle of IRSTD should be centroid-oriented detection instead of dense pixel-level reconstruction.}
Existing efforts to alleviate dense segmentation can be broadly categorized into two directions. \textit{(1) Bounding-box supervised regression frameworks.} These works adopt bounding box-based detection frameworks to simplify prediction~\cite{dai2023OSCAR,yang2024eflnet,REDETR_box,atrash2025tybox}; however, they still rely on bounding-box annotations and do not explicitly model the intrinsic local response structure of infrared small targets. \textit{(2) Single-point supervision-guided segmentation frameworks.} These works try to reduce annotation cost through point-level weak supervision with pseudo-mask generation~\cite{2023MCLC,2023LESPS,2025FromEasy2Hard,HE_SPS}, yet they continue to depend on encoder--decoder architectures and dense mask reconstruction. Although these approaches partially ease specific burdens, they do not simultaneously resolve the coupled bottlenecks of annotation efficiency, structural complexity, and localization stability.

Combining single-point supervision with an encoder-only architecture is conceptually simple, but deriving equally informative outputs from much sparser labels is non-trivial. We observe that infrared small targets contain limited global semantic information but remain locally salient within compact neighborhoods. Motivated by this observation, we propose a novel Single-Point Supervision guided Infrared Probabilistic Response Encoding method (namely, SPIRE), a probabilistic response modeling framework that reformulates IRSTD as spatial probability regression centered on target localization, rather than reconstructing dense pixel regions. SPIRE models structured local probabilistic responses consistent with infrared imaging physics and centroid-oriented detection objectives. Specifically, \textbf{(1) Single-point supervision.} SPIRE first combines single-point annotations with small-target characteristic priors, such as local saliency and Gaussian-like energy distribution. It then transforms sparse point labels into probabilistic response maps via Point-Response Prior Supervision (PRPS), reformulating target learning as structured local probability modeling. The supervision encodes a Gaussian-like spatial prior capturing center-enhanced and radially decaying energy responses while incorporating contrast-aware constraints to approximate realistic radiometric intensity transitions. Consequently, the network learns structured spatial probability distributions rather than discrete pixel labels, substantially increasing effective supervision density, alleviating gradient sparsity under extreme class imbalance, stabilizing optimization, and significantly reducing annotation cost.
\textbf{(2) Encoder-only and peak-extraction inference framework.} Once target discrimination is modeled through structured probabilistic responses rather than pixel-wise reconstruction, the necessity for heavy multi-scale fusion and attention-based compensation is substantially reduced. SPIRE adopts a High-Resolution Probabilistic Encoder (HRPE) that avoids aggressive down-sampling and cross-scale branches, thereby preserving the continuity and separability of local probabilistic responses. Built with cascaded lightweight channel-reorganization units, the encoder achieves expressive yet computationally efficient representation learning. Inference is performed by regressing probabilistic response maps and then applying fast deterministic peak extraction as lightweight post-processing, avoiding mask reconstruction and clustering-based post-processing.


The contributions of this paper can be summarized as:
\begin{itemize}
    \item Inspired by the intrinsic characteristics and engineering objectives of IRSTD, we rethink infrared small target detection as a centroid-oriented spatial probability regression task that can be learned using only point annotations, and accordingly establish a probabilistic response modeling paradigm termed SPIRE.
    
    \item We introduce a Point-Response Prior Supervision (PRPS) mechanism that converts sparse point labels into probabilistic response maps consistent with infrared imaging physics, inherently increasing supervision efficiency while avoiding pixel-level or box-level annotations.
    
    \item We design HRPE as a lightweight single-encoder module. It removes decoder branches and relies only on fast deterministic peak extraction as lightweight post-processing, enabling efficient centroid prediction.
    
    \item Extensive experiments demonstrate that SPIRE achieves competitive or superior detection accuracy with consistently low false alarms, while naturally maintaining lower annotation cost and computational complexity, as illustrated in \cref{fig:f1_pre_params_bubble}.
\end{itemize}

\section{Related Work}

\subsection{Infrared Small Target Detection}
Infrared small targets usually occupy only a few pixels and are often embedded in cluttered backgrounds with extremely low signal-to-noise ratio (SNR), which makes stable target representation and foreground–background discrimination particularly challenging. Early model-driven IRSTD methods \cite{arce1987theoretical2,bai2010analysis4,bai2018derivative3,chen2013local8,chen2014novel10,dai2017non12,deng2016small15,gao2013infrared19} addressed this problem by exploiting handcrafted priors such as local contrast enhancement, background suppression, and morphological filtering. With the emergence of annotated datasets, deep learning-based approaches \cite{2023DNANet29,2023LESPS,2023MCLC,2023UIUNet,2024ISPANet,2024MSHNet,2024STTrans,2025FromEasy2Hard,2025HDNet,2025L2SKNet,2025SIRST-UAVB_Dataset,2026FeedBack} have gradually become the dominant paradigm. CNN-based models commonly employ multi-scale feature fusion and dense cross-layer connections to preserve small target responses \cite{2023DNANet29,2023UIUNet,2024ISPANet,2024MSHNet,2025HDNet,2025L2SKNet}, while more recent works incorporate transformer modules and attention mechanisms to model long-range contextual dependencies and suppress background interference \cite{2024STTrans,2024SCTransNet}.
Despite these advances, most IRSTD approaches still rely on dense pixel-wise segmentation. Such formulations struggle with extreme target sparsity and often depend on heavy architectures and feature aggregation for stable detection.

\subsection{Single-Point Supervision}
To alleviate the high cost and ambiguity of pixel-wise annotation, recent studies introduce single-point supervision into IRSTD, where only target centroids are annotated. LESPS~\cite{2023LESPS} first explored this idea by generating pseudo masks from point annotations to supervise a segmentation network, reducing labeling cost while maintaining compatibility with existing architectures. Building upon this idea, MCLC~\cite{2023MCLC} further introduced multi-level constraints to refine pseudo-label generation and improve the quality of supervision derived from sparse annotations. More recently, FromEasy2Hard~\cite{2025FromEasy2Hard} proposed a curriculum-style learning strategy that progressively enhances supervision quality from coarse point guidance to refined target localization.
Although these methods reduce annotation effort, they largely remain within dense segmentation frameworks. Point annotations are typically converted into pseudo masks or auxiliary supervision, without fundamentally redefining detection as sparse centroid localization.

\subsection{Heatmap-Based Modeling}

Heatmap-based modeling is widely used in keypoint localization tasks such as human pose estimation and landmark detection \cite{2021HRNet,2020HigherHRnet,2024MTPose,2024SHaRPose}. 
These methods represent target locations as Gaussian heatmaps centered on annotated points and train networks to regress spatial confidence distributions. 
Representative approaches such as HRNet~\cite{2021HRNet} and HigherHRNet~\cite{2020HigherHRnet} achieve accurate localization by maintaining high-resolution representations or aggregating multi-scale heatmaps.
A few studies introduce heatmap supervision into IRSTD. 
For example, P2P-HDNet~\cite{2025P2PHDNet} formulates IRSTD as point-to-point heatmap prediction, but still relies heavily on feature fusion and attention-based compensation. 
More importantly, existing approaches rarely integrate the physical formation characteristics of infrared point targets, such as point spread function (PSF)-induced energy diffusion, into heatmap modeling or supervision design.

\section{Methodology}
\label{sec:method}

\begin{figure}[tb]
	\centering
	\includegraphics[width=\textwidth]{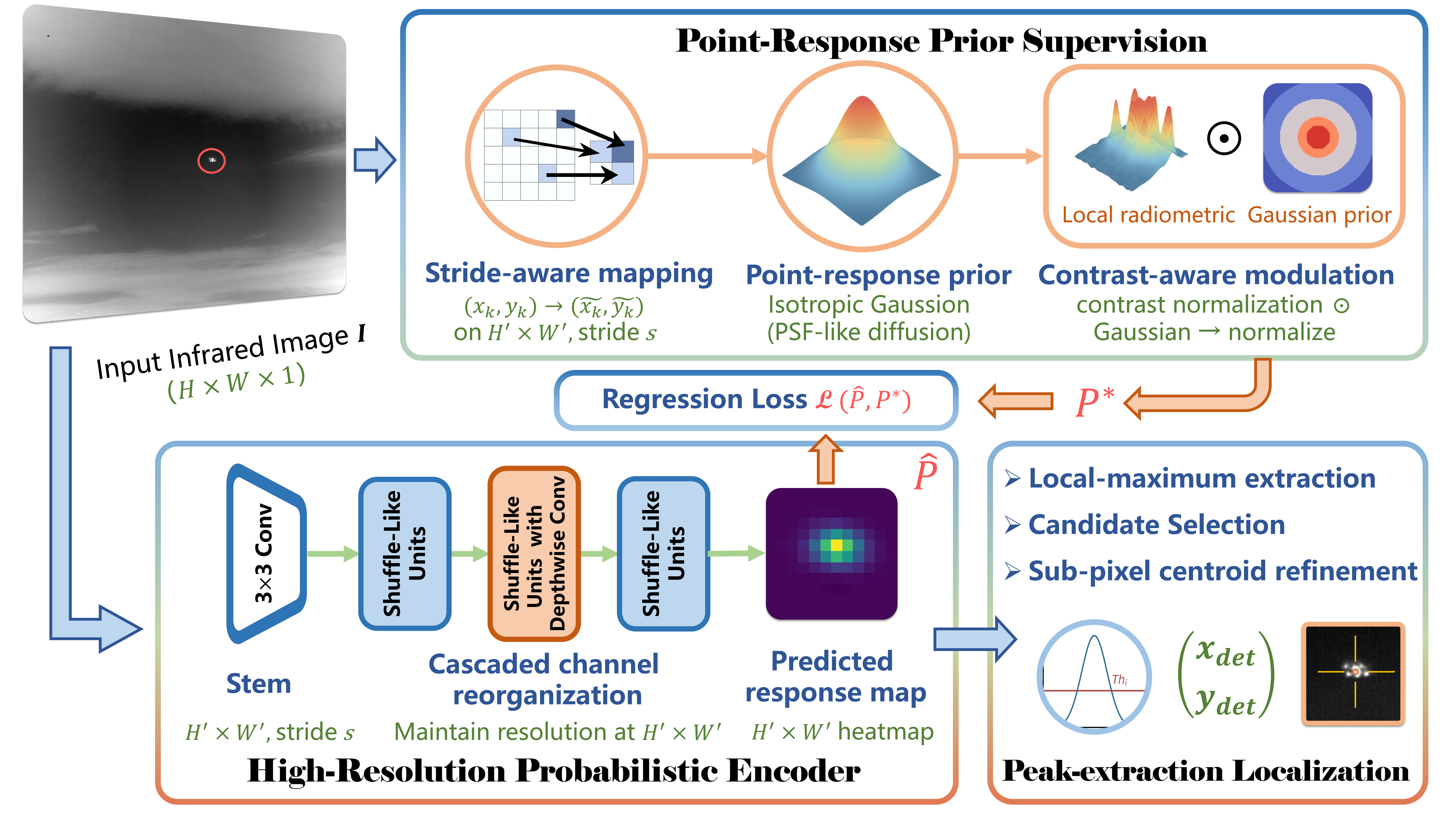}
	\caption{
\textbf{Framework of SPIRE (Single-Point Supervision guided Infrared Probabilistic Response Encoding).}
Given an infrared image with single-point centroid annotations, SPIRE constructs structured supervision through Point-Response Prior Supervision (PRPS), which maps centroids to the feature space and generates Gaussian-like response priors inspired by infrared PSF diffusion.
A lightweight High-Resolution Probabilistic Encoder (HRPE) then regresses the probabilistic response map (heatmap).
During inference, target locations are obtained by lightweight peak extraction on the predicted heatmap via local-maximum detection, thresholding, and sub-pixel centroid refinement.
}
	\label{fig:framework}
\end{figure}

\subsection{Overall Framework}
\label{sec:overall}

Infrared small targets occupy only a few pixels and exhibit Gaussian-like energy diffusion under the combined effects of the PSF and sensor sampling, while practical search and tracking systems primarily require reliable centroid localization rather than contour reconstruction. Conventional encoder--decoder segmentation paradigms are therefore structurally mismatched: dense pixel-wise supervision induces severe foreground--background imbalance and gradient sparsity, necessitating increasingly complex multi-scale fusion and attention compensation. 

As shown in \cref{fig:framework}, SPIRE reformulates IRSTD as centroid-oriented spatial probability regression, aligning both supervision and architecture with infrared imaging physics and detection objectives.
Given an input infrared image $\mathbf{I} \in \mathbb{R}^{H \times W \times 1}$, HRPE produces a single-channel probabilistic response map $\hat{\mathbf{P}} \in \mathbb{R}^{H' \times W'}$ at reduced resolution
\begin{equation}
	\label{eq:stride}
	H' = \lfloor H / s \rfloor, \quad W' = \lfloor W / s \rfloor,
\end{equation}
where $s$ denotes the output stride.

The ground-truth supervision map $\mathbf{P}^{*} \in \mathbb{R}^{H' \times W'}$ is constructed offline by PRPS from single-point annotations $\{(x_k, y_k)\}_{k=1}^{K}$, encoding structured local probabilistic responses consistent with infrared point-target response characteristics (\cref{sec:PRPS}). Target centroid locations are obtained via a lightweight deterministic peak-extraction procedure on $\hat{\mathbf{P}}$ (\cref{sec:inference}), rather than mask reconstruction or iterative post-processing.

\subsection{Point-Response Prior Supervision}
\label{sec:PRPS}

Under PSF-induced blur and imaging degradation, infrared small targets exhibit locally concentrated responses with radially decaying intensity rather than isolated impulses.
PRPS encodes a point-response prior together with observed local radiometric structure, generating structured local probabilistic responses that capture isotropic diffusion and contrast-aware radiometric transitions.

\textbf{Stride-aware coordinate mapping.}
Each annotated centroid $(x_k, y_k)$ is mapped to the heatmap lattice via
\begin{equation}
\label{eq:coord}
(\tilde{x}_k, \tilde{y}_k)
=
\left\lfloor
\left(\frac{x_k}{s},\, \frac{y_k}{s}\right) + 0.5
\right\rfloor,
\end{equation}
where $s$ denotes the output stride.
The centroid is further refined to the position of maximum intensity within a local neighborhood in the original image, yielding $(\tilde{x}_k^{*}, \tilde{y}_k^{*})$ aligned with the true radiometric peak.

\textbf{Gaussian spatial prior.}
An isotropic Gaussian kernel centered at $(\tilde{x}_k^{*}, \tilde{y}_k^{*})$ encodes PSF-induced energy diffusion:
\begin{equation}
\label{eq:gauss}
G_k(u,v)
=
\exp\!\left(
-\frac{(u-\tilde{x}_k^{*})^2+(v-\tilde{y}_k^{*})^2}{2\sigma^2}
\right),
\quad
\|(u,v)-(\tilde{x}_k^{*},\tilde{y}_k^{*})\|_\infty \le r,
\end{equation}
where $\sigma$ controls spatial spread and $r=3\sigma$ defines the compact support radius beyond which the kernel is truncated to zero.
The kernel has unit peak and is not normalized.
PRPS is not intended to reconstruct target contours; instead, it provides a centroid-oriented local probabilistic response. We set $\sigma=2$ and $r=6$ in all experiments. This scale choice is empirically examined in Sec.~\ref{sec:ablation} and Tab.~\ref{tab:ab_PRPS_uavb}.

\textbf{Contrast-aware modulation.}
A $(2r+1)\times(2r+1)$ patch centered at $(\tilde{x}_k^{*}, \tilde{y}_k^{*})$ is extracted from the original image and contrast-normalized:
\begin{equation}
\label{eq:contrast}
C_k(u,v)
=
\frac{\mathbf{I}_{\mathrm{patch}}(u,v)-\mathbf{I}_{\mathrm{patch}}^{\min}}
{\mathbf{I}_{\mathrm{patch}}^{\max}-\mathbf{I}_{\mathrm{patch}}^{\min}},
\end{equation}
where $\mathbf{I}_{\mathrm{patch}}^{\min}$ and $\mathbf{I}_{\mathrm{patch}}^{\max}$ denote the minimum and maximum intensities within the patch.
The per-target response is then composed as
\begin{equation}
\label{eq:compose}
H_k = \mathrm{Norm}(G_k \odot C_k),
\end{equation}
where $\odot$ denotes element-wise multiplication and $\mathrm{Norm}(\cdot)$ performs min-max normalization to $[0,1]$.
This preserves the center-enhanced diffusion pattern while adapting to the observed local radiometric contrast.

\textbf{Multi-target aggregation.}
For images containing $K$ targets, responses are aggregated into a single supervision map:
\begin{equation}
\label{eq:agg}
\mathbf{P}^*(u,v)
=
\mathcal{A}\!\left(\{H_k(u,v)\}_{k=1}^{K}\right),
\end{equation}
where $\mathcal{A}$ denotes deterministic aggregation.
Even when Gaussian supports overlap, centroid-oriented regression preserves distinct local maxima for each target, enabling deterministic peak extraction without mask clustering.
Overlapping supports therefore do not necessarily merge nearby targets when separable local maxima are preserved.
By expanding sparse point labels into spatially structured responses consistent with infrared point-target response characteristics, PRPS increases supervision density and alleviates gradient starvation under extreme foreground--background imbalance.

\textbf{Training objective.}
The network is optimized via standard pixel-wise regression between $\hat{\mathbf{P}}$ and $\mathbf{P}^*$.
Because PRPS provides dense structured supervision within each kernel support, sufficient gradient contributions are maintained to ensure stable convergence without specialized imbalance-aware losses.

\subsection{High-Resolution Probabilistic Encoder}
\label{sec:hrpe}

Once target discrimination is grounded in structured local probabilistic responses rather than pixel-wise mask reconstruction, heavy multi-scale fusion and decoder-based compensation become unnecessary.
HRPE is a purely encoder-only, single-branch architecture that maintains high-resolution representations at a constant output stride $s$.
This design is critical for IRSTD, as aggressive downsampling may destroy the few-pixel support of dim targets and erase sub-pixel centroid cues under low signal-to-noise ratio (SNR) and low signal-to-clutter ratio (SCR) conditions.

\textbf{Stem.}
Two successive $3{\times}3$ convolutions with stride 2, each followed by batch normalization and ReLU, reduce the resolution to $H/s \times W/s$ while expanding channels to $C_0{=}64$.
A bottleneck residual block and lightweight channel reorganization units further refine features at the same resolution.

\textbf{Cascaded channel reorganization.}
After a transition layer reduces channels to $C{=}32$, features pass through cascaded lightweight channel reorganization units operating on a single-resolution branch.
Given input $\mathbf{F} \in \mathbb{R}^{C \times H' \times W'}$, features are evenly split into $\mathbf{F}_a, \mathbf{F}_b \in \mathbb{R}^{C/2 \times H' \times W'}$.
One branch is preserved as identity, while the other undergoes depthwise separable convolution and channel-wise attention reweighting:

\begin{equation}
\mathbf{F}' =
\mathrm{Shuffle}\!\Big(
\mathrm{Cat}\big(
\mathbf{F}_a,\;
\phi_{\mathrm{attn}}\!\left(
\mathrm{DWConv}(\mathbf{F}_b)
\right)
\big)
\Big),
\end{equation}
where $\mathrm{DWConv}$ denotes $3{\times}3$ depthwise convolution with batch normalization, $\phi_{\mathrm{attn}}$ applies squeeze-and-excitation-style reweighting, and $\mathrm{Shuffle}$ interleaves channels for cross-group information exchange.
Some units further include an additional depthwise convolution for local spatial refinement.
This split--process--shuffle design~\cite{zhang2018shufflenet} enables expressive yet efficient channel reorganization while preserving spatial continuity and separability for structured probabilistic response learning.

No multi-resolution branches, decoder pathways, or skip connections are used, yielding a strictly encoder-only architecture aligned with centroid-oriented probabilistic regression.

\textbf{Prediction head.}
A single $1{\times}1$ convolution maps the $C$-channel representation to one output channel, producing $\hat{\mathbf{P}} \in \mathbb{R}^{H' \times W'}$.
No activation is applied; outputs are directly regressed to the PRPS target.

\subsection{Lightweight Probabilistic Inference}
\label{sec:inference}

At inference time, target centroid locations are extracted from $\hat{\mathbf{P}}$ using fast deterministic peak extraction as lightweight post-processing, rather than mask reconstruction, connected-component analysis, or learned post-processing.

\textbf{Local-maximum extraction.}
Following standard practice~\cite{2021HRNet}, a $3{\times}3$ max-pooling non-maximum suppression (NMS) with stride 1 suppresses non-maximum responses:
\begin{equation}
\label{eq:nms}
\hat{\mathbf{P}}_{\mathrm{nms}}(u,v)
=
\begin{cases}
\hat{\mathbf{P}}(u,v), &
\hat{\mathbf{P}}(u,v)
=
\max\limits_{(u',v') \in \mathcal{N}_{3 \times 3}(u,v)}
\hat{\mathbf{P}}(u',v'),\\
0, & \mathrm{otherwise}.
\end{cases}
\end{equation}

\textbf{Thresholding and candidate selection.}
We first collect NMS responses whose confidence exceeds $\tau$:
\begin{equation}
\label{eq:peakset}
\mathcal{M}_{\tau}
=
\big\{(u,v)\mid \hat{\mathbf{P}}_{\mathrm{nms}}(u,v)>\tau\big\}.
\end{equation}
Here $\mathcal{M}_{\tau}$ denotes the thresholded local-maximum set, $\mathcal{N}_{3\times3}(u,v)$ is the NMS neighborhood, and $\tau$ is the confidence threshold. Experiments show that $\tau$ has a broad stable range and does not require dataset-specific retuning.

Final detections are then selected by response-ranked top-$N_{\max}$:
\begin{equation}
\label{eq:detect}
\mathcal{D}
=
\mathrm{top}\text{-}N_{\max}(\mathcal{M}_{\tau};\hat{\mathbf{P}}_{\mathrm{nms}}).
\end{equation}
Here $\mathcal{D}$ is the predicted centroid set, and $\mathrm{top}\text{-}N_{\max}$ returns at most $N_{\max}$ peaks with the highest NMS response scores. We set $N_{\max}$ to the maximum target count of each dataset; increasing it to several times that count leaves the metrics unchanged because extra low-score peaks are filtered by $\tau$.

\textbf{Sub-pixel centroid refinement.}
Each integer-coordinate peak $(u_0, v_0) \in \mathcal{D}$ is refined using the sign of the local finite-difference gradient on the pre-NMS response map~\cite{2021HRNet}:
\begin{equation}
\label{eq:subpix}
\begin{aligned}
\Delta u
&=
\tfrac{1}{s}\,
\mathrm{sign}\!\big(
\hat{\mathbf{P}}(u_0{+}1,\,v_0)
-
\hat{\mathbf{P}}(u_0{-}1,\,v_0)
\big),
\\
\Delta v
&=
\tfrac{1}{s}\,
\mathrm{sign}\!\big(
\hat{\mathbf{P}}(u_0,\,v_0{+}1)
-
\hat{\mathbf{P}}(u_0,\,v_0{-}1)
\big).
\end{aligned}
\end{equation}
Here $s$ denotes the output stride. In this work we set $s=4$, as validated in \cref{sec:ablation}, achieving a favorable balance between spatial fidelity and computational efficiency.
This finite-difference refinement follows standard heatmap localization and serves as a lightweight deterministic correction.

\textbf{Coordinate back-mapping.}
Refined probabilistic response map coordinates are mapped to the original image space via the inverse affine transformation:
\begin{equation}
\label{eq:backmap}
\begin{pmatrix}
x_{\mathrm{det}} \\
y_{\mathrm{det}}
\end{pmatrix}
=
\mathbf{T}^{-1}
\begin{pmatrix}
u_0 + \Delta u \\
v_0 + \Delta v \\
1
\end{pmatrix},
\end{equation}
where $\mathbf{T}^{-1} \in \mathbb{R}^{2 \times 3}$ restores coordinates from the stride-reduced probabilistic response map to the original infrared image grid, preserving geometric consistency between centroid-oriented spatial probability regression and the physical image coordinate system.

The complete procedure---a single forward pass through HRPE followed by fast deterministic peak extraction as lightweight post-processing---avoids mask reconstruction and iterative segmentation post-processing, directly yielding target centroid coordinates consistent with the centroid-oriented spatial probability regression formulation of SPIRE.

\section{Experiment}
\subsection{Experimental Setup}
\label{sec:exp_setup}

\textbf{Datasets.}
Experiments are conducted on two public benchmarks, SIRST-UAVB and SIRST4~\cite{2025SIRST-UAVB_Dataset}. SIRST-UAVB contains 3000 images and SIRST4 contains 3352 images; their train/test splits are 2400/600 and 2285/1067, respectively.
SIRST4 integrates SIRSTv1~\cite{dai2021acm13}, SIRSTv2~\cite{dai2023OSCAR}, NUDT-SIRST~\cite{2023DNANet29}, and IRSTD-1K~\cite{zhang2022isnet68}, covering diverse scenes and target scales. SIRST-UAVB focuses on UAV-based urban scenes with clutter and low SNR.
Both datasets include real and simulated targets, isolated and clustered targets, as well as empty-background images. 
Mask-level and point-level annotations are provided; we use centroid point annotations for supervision and retain mask annotations for fair comparison.

\textbf{Implementation details.}
All models are implemented in PyTorch and trained on an RTX 4090 GPU. 
We adopt Adam with an initial learning rate of 0.005 and a ReduceLROnPlateau scheduler (factor 0.01, patience 3). 
The batch size is 10. 
Each competing method follows its original training schedule, while our model is trained for 500 epochs. 
Input resolutions are $512 \times 512$ for SIRST4 and $640 \times 640$ for SIRST-UAVB.

\textbf{Evaluation metrics.}
IRSTD is evaluated at the centroid level. 
Segmentation outputs are converted to centroids via 8-connected clustering \cite{2023DNANet29}, bounding-box detectors use box centers, and probabilistic response map-based methods use peak extraction. 
A prediction is counted as a true positive (TP) if its Euclidean distance to a ground-truth centroid is within $\delta=5$ pixels; otherwise it is a false positive (FP). 
We use this threshold as the default protocol and later verify the robustness to $\delta$ in Sec.~\ref{sec:ablation} and Table~\ref{tab:multi_delta_sirst4}.
Unmatched ground-truth targets are counted as false negatives (FN). 

We report Precision, Recall, and F1-score, defined as
\begin{align}
	\mathrm{Precision} &= \frac{\mathrm{TP}}{\mathrm{TP} + \mathrm{FP}}, \\
	\mathrm{Recall} &= \frac{\mathrm{TP}}{\mathrm{TP} + \mathrm{FN}},
\end{align}
where Recall is equivalent to the probability of detection (Pd) in IRSTD. 
The F1-score is
\begin{equation}
	\mathrm{F1} =
	\frac{2 \cdot \mathrm{Precision} \cdot \mathrm{Recall}}
	{\mathrm{Precision} + \mathrm{Recall}}.
\end{equation}

The false alarm rate (Fa) is defined as
\begin{equation}
	\mathrm{Fa} = \frac{\mathrm{FP}}{N_{\mathrm{pixels}}},
\end{equation}
where $N_{\mathrm{pixels}}$ denotes the total number of image pixels. 
We additionally report floating-point operations (FLOPs) and the number of parameters (Params) for efficiency comparison.

\subsection{Comparison with State of the Art}
\label{sec:sota}

\begin{figure}[tb]
	\centering
	\includegraphics[width=\textwidth]{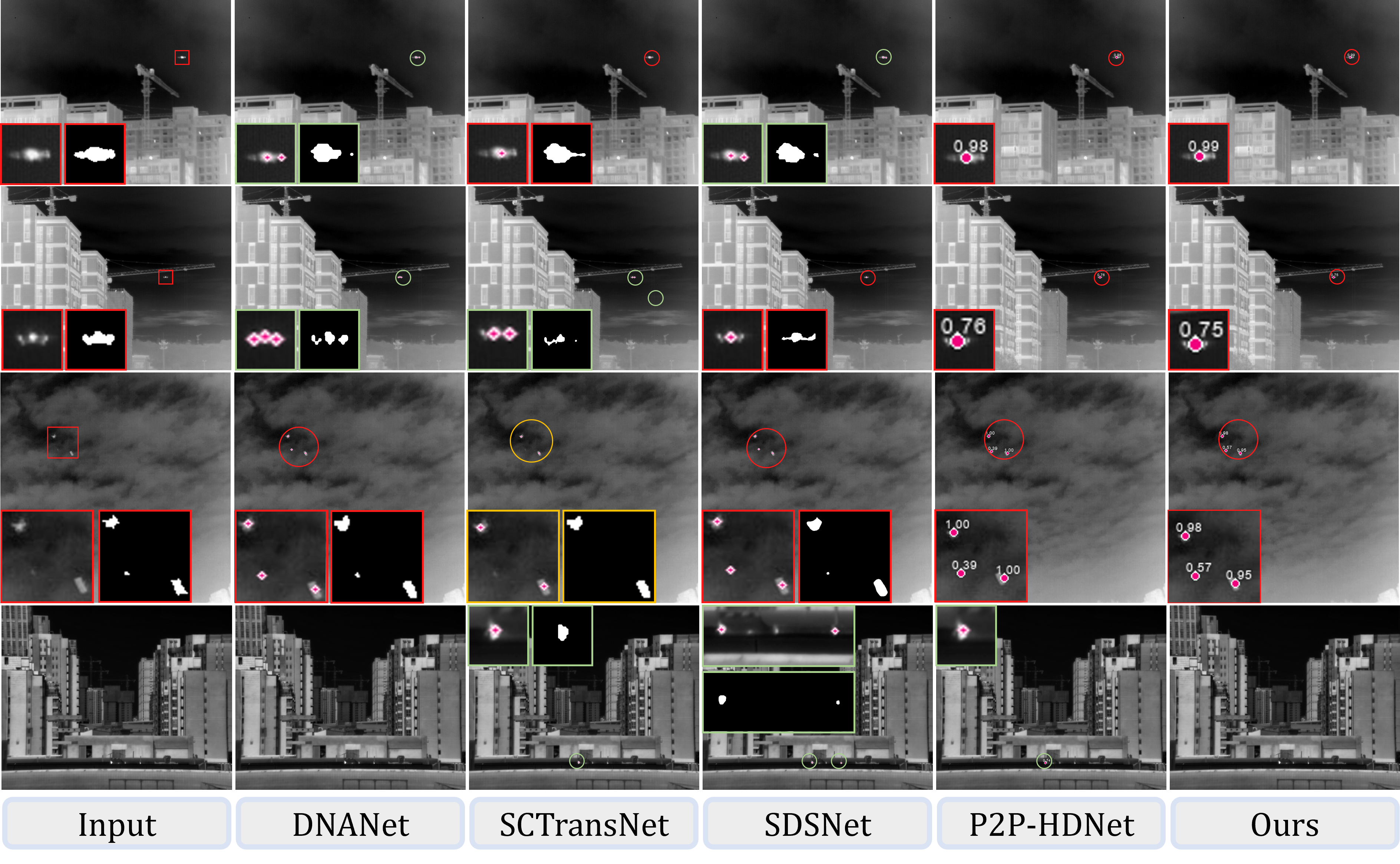}
	\caption{
\textbf{Qualitative comparison of SPIRE on SIRST-UAVB and SIRST4.}
Red rectangles denote ground-truth targets.
Red circles indicate true positives, green circles denote false positives, and orange circles correspond to false negatives under the unified centroid protocol.
}
	\label{fig_expIMG}
\end{figure}

\cref{tab:quant_sirst} reports quantitative results on SIRST-UAVB and SIRST4, and \cref{fig_expIMG} provides qualitative comparisons. 

\textbf{Quantitative comparison.}
Table~\ref{tab:quant_sirst} reports quantitative results under the unified centroid-level matching protocol. On SIRST-UAVB, our method achieves the highest Precision of 99.82 and the lowest Fa of 1.02, while maintaining F1 of 97.05. Compared with DNANet and SCTransNet, which report Fa of 15.77 and 5.09 respectively, our method consistently suppresses false alarms while using substantially fewer parameters. 
On SIRST4, our method achieves F1 of 94.60 while requiring over 8 times fewer FLOPs than SCTransNet and more than 16 times fewer parameters than DNANet. These results demonstrate that centroid-oriented spatial probability regression achieves competitive detection accuracy while naturally decoupling structural complexity from performance within an encoder-only framework.
To complement the FLOPs and Params in Table~\ref{tab:quant_sirst}, we further benchmark frames per second (FPS) for representative high-accuracy competitors, as summarized in Table~\ref{tab:multi_delta_sirst4}. SPIRE reaches 261.2 FPS, running $3.23{\times}$ faster than DNANet, $3.89{\times}$ faster than SCTransNet, and $5.56{\times}$ faster than SDSNet. This confirms that the encoder-only design improves practical throughput while maintaining competitive detection accuracy.

\textbf{Qualitative comparison.}
Figure~\ref{fig_expIMG} visualizes representative detection results. Red rectangles denote ground-truth targets. Red circles indicate true positives, green circles denote false positives, and orange circles correspond to false negatives under the unified centroid protocol. Segmentation-based methods such as DNANet and SCTransNet often produce multiple adjacent mask responses around a single target. After connected-component clustering, these redundant responses may generate extra centroids near true targets, increasing false alarms. In contrast, our structured probabilistic response regression produces a single dominant peak per target without neighboring over-activation, resulting in cleaner centroid localization and consistently lower Fa.

Overall, both quantitative and qualitative evidence validate that sparse centroid modeling with physically grounded probabilistic supervision provides a more efficient and principled alternative to dense mask reconstruction for IRSTD.

\begin{table*}[t]
	\caption{\textbf{Comparison with representative methods on SIRST-UAVB and SIRST4.} We report Pre(\%), Rec (Pd)(\%), F1(\%), Fa ($10^{-8}$), Params (M), and FLOPs (G). 
Params and FLOPs are computed at $640 \times 640$ input resolution.}
	\label{tab:quant_sirst}
	\centering
	\setlength{\tabcolsep}{3pt}
	
	\resizebox{\textwidth}{!}{%
		\begin{tabular}{
				l
				c
				*{10}{>{\centering\arraybackslash}p{1.15cm}}
			}
			\toprule
			
			\multirow{2}{*}{\textbf{Method}} &
			\multirow{2}{*}{\textbf{Venue}} &
			\multicolumn{4}{c}{\textbf{SIRST-UAVB} (2400:600)} &
			\multicolumn{4}{c}{\textbf{SIRST4} (2285:1067)} &
			\multirow[c]{2}{*}{\shortstack[c]{\textbf{FLOPs}}$\downarrow$}
			&
			\multirow[c]{2}{*}{\shortstack[c]{\textbf{Params}}$\downarrow$} \\
			
			\cmidrule(lr){3-6}\cmidrule(lr){7-10}
			
			& & \textbf{Pre} $\uparrow$ & \textbf{Rec} $\uparrow$& \textbf{F1} $\uparrow$& \textbf{Fa} $\downarrow$
			& \textbf{Pre} $\uparrow$& \textbf{Rec} $\uparrow$& \textbf{F1} $\uparrow$& \textbf{Fa} $\downarrow$
			& & \\
			
			\midrule
			ACM \cite{dai2021acm13}        & $\mathrm{WACV}^{21}$ & 87.01 & 71.16 & 78.29 & 34.04  & 90.17 & 70.38 & 79.05 & 44.08  & \underline{2.51} & 0.40 \\
			ALCNet \cite{dai2021alcnet14}     & $\mathrm{TGRS}^{21}$ & 95.06 & 81.11 & 87.53 & 12.72  & 91.45 & 72.41 & 80.82 & 38.90  & \textbf{2.36} & 0.43 \\
			ISTDU-Net \cite{ISTDU-Net}   & $\mathrm{GRSL}^{22}$ & 79.28 & 78.08 & 78.67 & 61.54  & \underline{93.28} & 72.03 & 81.29 & 29.83  & 49.65 & 2.75 \\
			RDIAN \cite{RDIAN}      & $\mathrm{TGRS}^{23}$ & 52.68 & 77.91 & 62.86 & 211.08 & 90.64 & 80.83 & 85.45 & 47.98  & 23.24 &\textbf{ 0.22} \\
			DNANet \cite{2023DNANet29}    & $\mathrm{TIP}^{23}$  & 94.48 & 89.54 & 91.95 & 15.77  & 93.99 & 81.20 & 87.13 & \underline{29.82}  & 89.13 & 4.70 \\
			SCTransNet \cite{2024SCTransNet}& $\mathrm{TGRS}^{24}$ & \underline{98.27} &\underline{95.95} & \textbf{97.09} & \underline{5.09}   & 81.20 & 86.39 & 83.72 & 114.98 & 63.22 & 11.19 \\
			MSHNet \cite{2024MSHNet}    & $\mathrm{CVPR}^{24}$ & 71.90 & 88.02 & 79.09 & 104.27 & 90.68 & \underline{92.86} & 91.47 & 60.51  & 38.16 & 4.07 \\
			SDSNet \cite{2025SDSNet}    & $\mathrm{TGRS}^{25}$ & 97.60 & \textbf{96.29} & 96.94 & 7.12   & 87.35 & 88.27 & 87.81 & 73.48  & 42.42 & 2.49 \\
			L$^{2}$SKNet  \cite{2025L2SKNet}  & $\mathrm{TGRS}^{25}$ & 95.83 & 89.04 & 92.31 & 11.70  & 92.89 & 91.42 & \underline{92.16} & 40.20  & 43.09 & 0.90 \\
			\textbf{Ours}    & -                   & \textbf{99.82} & 94.44 & \underline{97.05} & \textbf{1.02}
			& \textbf{95.00} & \textbf{94.21} & \textbf{94.60} & \textbf{28.53}
			& 7.68 & \underline{0.29} \\
			
			\bottomrule
		\end{tabular}%
	}
\end{table*}

\subsection{Ablation and Analysis}
\label{sec:ablation}

We first assess the robustness of the evaluation protocol, and then analyze the key design choices of SPIRE.

\noindent\textbf{Robustness to matching threshold $\delta$.}
Since centroid-level evaluation depends on the matching tolerance, we examine whether the target-level conclusion is sensitive to this setting. On SIRST4, Table~\ref{tab:multi_delta_sirst4} shows that the stricter $\delta=3$ setting reduces Recall for all methods, as the tolerance becomes comparable to the average target radius of 3.31 pixels. SPIRE remains most stable in F1 (94.74 vs. 94.60 under $\delta=5,8,10$). The identical results for $\delta=5,8,10$ indicate that increasing the tolerance beyond 5 does not change the evaluated matches, validating $\delta=5$ as a reasonable default.

\begin{table}[t]
	\caption{\textbf{Multi-threshold robustness and inference speed comparison on SIRST4.} We report Rec (Pd)(\%), F1(\%), Fa ($10^{-8}$) under centroid-matching thresholds, together with FPS measured under the same input and hardware setting.}
	\label{tab:multi_delta_sirst4}
	\centering
	\setlength{\tabcolsep}{4.5pt}
	\begin{tabular}{lcccccc||c}
		\toprule
		\multicolumn{1}{c}{\multirow{2}{*}{\textbf{Method}}} &
		\multicolumn{3}{c}{$\delta=3$} &
		\multicolumn{3}{c||}{$\delta=5,8,10$} &
		\multirow{2}{*}{\textbf{FPS} $\uparrow$} \\
		\cmidrule(lr){2-4}\cmidrule(lr){5-7}
		& Rec $\uparrow$ & F1 $\uparrow$ & Fa $\downarrow$
		& Rec $\uparrow$ & F1 $\uparrow$ & Fa $\downarrow$
		& \\
		\midrule
		DNANet\cite{2023DNANet29} & 80.52 & 86.41 & 33.72 & 81.20 & 87.13 & 29.82 & 80.86 \\
		SCTransNet\cite{2024SCTransNet} & 80.15 & 81.18 & 99.42 & 86.39 & 83.72 & 114.98 & 67.09 \\
		SDSNet\cite{2025SDSNet} & 86.39 & 87.21 & 67.43 & 88.27 & 87.81 & 73.48 & 46.97 \\
		\textbf{Ours} & \textbf{92.93} & \textbf{94.74} & \textbf{18.61} & \textbf{94.21} & \textbf{94.60} & \textbf{28.53} & \textbf{261.2} \\
		\bottomrule
	\end{tabular}
\end{table}



\begin{table}[t]
	\caption{\textbf{Ablation study of PRPS on SIRST-UAVB.}
		The first block compares different supervision forms, while the second block analyzes the scale sensitivity of PRPS. We report Pre(\%), Rec (Pd)(\%), F1(\%), and Fa ($10^{-8}$)}
	\label{tab:ab_PRPS_uavb}
	\centering
	\setlength{\tabcolsep}{4.5pt}
	\begin{tabular}{lcccc}
		\toprule
		\textbf{Variant} & Pre $\uparrow$ & Rec $\uparrow$ & $F_1$ $\uparrow$ & Fa $\downarrow$ \\
		\midrule
		
		\multicolumn{5}{c}{\textit{Supervision form}} \\
		\midrule
		\textbf{PRPS} & \textbf{99.82} & \textbf{94.44}& \textbf{97.05} & \textbf{1.02} \\
		Single-point impulse & 98.71 & 90.22 & 94.27 & 2.85 \\
		Unconstrained Gaussian & 98.93 & 93.76 & 96.28 & 2.44 \\
		
		\midrule
		\multicolumn{5}{c}{\textit{PRPS scale sensitivity}} \\
		\midrule
		\textbf{PRPS ($\sigma{=}2,\ r{=}6$)} & \textbf{99.82} & \textbf{94.44}& \textbf{97.05} & \textbf{1.02} \\
		PRPS ($\sigma{=}1,\ r{=}3$) & 97.60 & 88.60 & 92.87 & 5.26 \\
		PRPS ($\sigma{=}3,\ r{=}9$) & 99.70 & 92.50 & 95.99 & 0.67 \\
		
		\bottomrule
	\end{tabular}
\end{table}
\begin{table}[t]
\caption{\textbf{Ablation study of HRPE on SIRST-UAVB.}
The first block evaluates resolution preservation with different output strides $s$,
while the second block analyzes the contribution of HRPE components. We report Pre(\%), Rec (Pd)(\%), F1(\%), Fa ($10^{-8}$), Params (\textbf{$10^{-2}$}M), and FLOPs (G). 
Params and FLOPs are computed at $640 \times 640$ input resolution.}
\label{tab:ab_hrpe_uavb}
\centering
\setlength{\tabcolsep}{4.5pt}
\begin{tabular}{
l
cccc
>{\centering\arraybackslash}p{1cm}
>{\centering\arraybackslash}p{1cm}
}
\toprule
\textbf{Variant} & Pre $\uparrow$ & Rec $\uparrow$ & $F_1$ $\uparrow$ & Fa $\downarrow$ & FLOPs$\downarrow$ & Params$\downarrow$\\
\midrule

\multicolumn{7}{c}{\textit{Resolution preservation (output stride)}}\\
\midrule
\textbf{HRPE ($s{=}4$)} &\textbf{99.82} & \textbf{94.44}& \textbf{97.05} & 1.02 & 7.68 & 29.47\\
HRPE ($s{=}2$) & 99.63 & 91.23 & 95.25 & \textbf{0.81} & 26.77 & 26.56\\
HRPE ($s{=}8$) & 90.05 & 83.98 & 86.91 & 22.38 & \textbf{3.05} & 33.27\\

\midrule
\multicolumn{7}{c}{\textit{Component ablation}}\\
\midrule
\textbf{HRPE} & \textbf{99.82} & \textbf{94.44}& \textbf{97.05} & \textbf{1.02} & \textbf{7.68} & \textbf{29.47}\\
w/o channel reorganization & 97.70 & 93.90 & 95.76 & 5.33 & 7.70 & 29.45\\
w/o reweighting & 96.52 & 93.71 & 95.09 & 8.15 & 7.72 & 29.63\\

\bottomrule
\end{tabular}
\end{table}

\begin{figure}[tb]
	\centering
	\includegraphics[width=\textwidth]{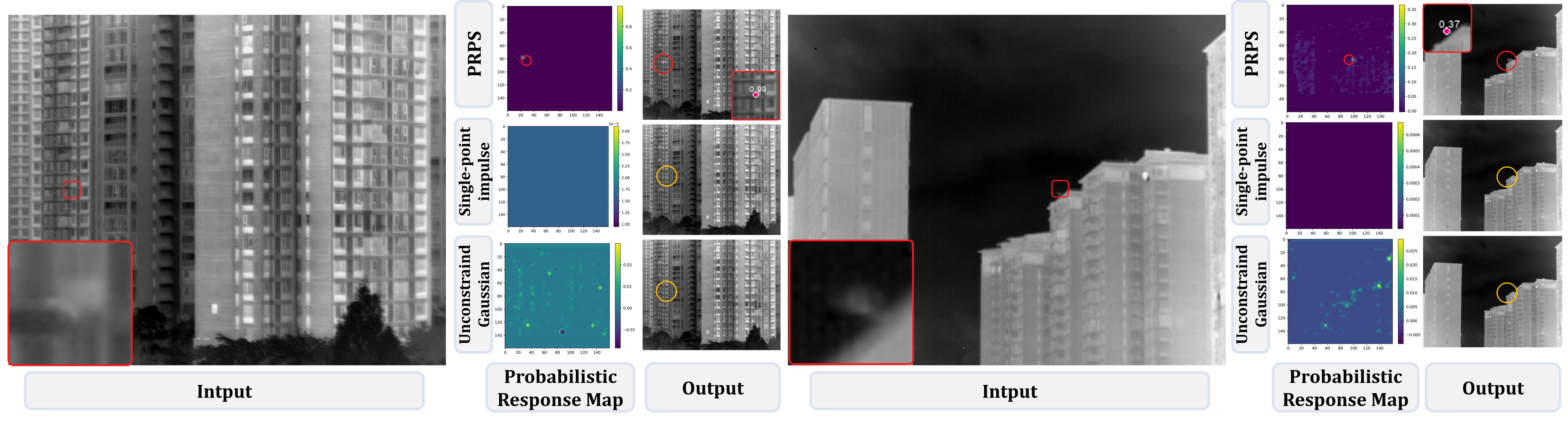}
	\caption{
\textbf{Ablation study of PRPS on SIRST-UAVB.}
Red rectangles denote ground-truth targets.
Red circles indicate true positives and orange circles correspond to false negatives under the unified centroid protocol.
}
	\label{fig:abprps}
\end{figure}

\noindent\textbf{Impact of PRPS supervision form.}
We compare PRPS with single-point impulse and unconstrained Gaussian supervision. As shown in Tab.~\ref{tab:ab_PRPS_uavb}, impulse supervision decreases Recall from 94.44 to 90.22 and increases Fa from 1.02 to 2.85, reducing F1 from 97.05 to 94.27. Gaussian supervision improves Recall to 93.76 but still yields higher Fa of 2.44 and lower F1 than PRPS. As shown in \cref{fig:abprps}, impulse supervision suffers from gradient starvation under extreme foreground–background imbalance, while isotropic Gaussian lacks the response-adaptive modulation guided by local radiometric structure, leading to less discriminative peaks in clutter. By encoding Gaussian responses consistent with infrared point-target characteristics, PRPS enables stable centroid-oriented spatial probability regression and suppresses spurious peaks.

\noindent\textbf{Impact of PRPS scale.}
We vary $\sigma$ to analyze scale sensitivity. As shown in Tab.~\ref{tab:ab_PRPS_uavb}, $\sigma = 2$ achieves the best F1 of 97.05. When $\sigma = 1$, Recall drops to 88.60 and Fa rises to 5.26; when $\sigma = 3$, Fa decreases to 0.67 but Recall declines to 92.50. A small $\sigma$ provides insufficient supervision density, while a large $\sigma$ over-smooths probability peaks and weakens centroid separability. The moderate scale best matches the intrinsic spatial extent of infrared small targets and yields balanced probabilistic regression. This trend further clarifies that PRPS is not intended to reconstruct target contours; instead, it should provide a compact, centroid-oriented local probabilistic response for localization.

\noindent\textbf{Impact of resolution preservation.}
We vary the output stride to evaluate high-resolution representation in HRPE. As shown in Tab.~\ref{tab:ab_hrpe_uavb}, stride 8 reduces Recall to 83.98 and increases Fa to 22.38, severely degrading F1. Stride 2 slightly lowers Fa to 0.81 but does not improve F1 and increases FLOPs from 7.68 to 26.77. Excessive downsampling weakens fragile small-target representations under low-SNR conditions, undermining structured probabilistic response modeling. Overly fine resolution increases structural redundancy without clear discriminative gain. Stride 4 provides the best balance within the encoder-only framework.

\noindent\textbf{Impact of HRPE components.}
Removing channel reorganization increases Fa to 5.33 and reduces F1 to 95.76, while removing reweighting further increases Fa to 8.15 with slight Recall degradation, as shown in Tab.~\ref{tab:ab_hrpe_uavb}. These results indicate that efficient channel interaction and adaptive refinement are essential for structured probabilistic response learning in clutter. Removing either component primarily weakens false-alarm control.


\section{Conclusion}
\label{sec:conclusion}

We revisit IRSTD from a modeling perspective and argue that dense pixel-wise segmentation is not inherently required for reliable centroid localization. 
By reformulating IRSTD as spatial probability regression, we propose SPIRE, which integrates PRPS for structured probabilistic supervision guided by point-response priors under extreme class imbalance and a lightweight HRPE for encoder-only probabilistic response map regression with lightweight peak-based inference rather than mask reconstruction.
Experiments on SIRST-UAVB and SIRST4 demonstrate that SPIRE achieves competitive target-level detection performance with consistently low Fa and significantly reduced FLOPs and Params. 
These results validate that sparse centroid modeling with physically grounded probabilistic supervision provides an efficient and principled alternative to dense mask reconstruction paradigms for IRSTD.

\section*{Acknowledgements}
This work was partially supported in part by the National Natural Science Foundation of China (NSFC) under No. 62101567, No.62401589, No. 62501618 and No. 4250013163.

%
%
\bibliographystyle{splncs04}
\bibliography{main}
\end{document}